\pdfoutput=1
\documentclass[11pt]{article}

\usepackage[]{ACL2023}

\usepackage{times}
\usepackage{latexsym}

\usepackage[T1]{fontenc}
\usepackage[utf8]{inputenc}

\usepackage{microtype}

\usepackage{inconsolata}

\usepackage{geometry}
\usepackage{graphicx}
\usepackage{booktabs}
\usepackage{multirow}
\usepackage{xspace}
\usepackage{amsmath}
\usepackage{amsfonts}
\usepackage{amssymb}
\usepackage{cleveref}
\usepackage{subcaption}
\usepackage{xcolor}
\usepackage{multicol}
\usepackage{bbm}

\title{Model-Generated Pretraining Signals Improves \\
Zero-Shot Generalization of Text-to-Text Transformers}

\definecolor{cxgreen}{RGB}{18, 170, 156}

\definecolor{tab10blue}{HTML}{1F77B4}
\definecolor{tab10orange}{HTML}{FF7F0E}
\definecolor{tab10green}{HTML}{2CA02C}
\definecolor{tab10red}{HTML}{D62728}

\author{
Linyuan Gong\textsuperscript{\normalfont 1}\thanks{\ \ Part of this work is done during Linyuan and Yiqing's internship at Microsoft.}, Chenyan Xiong\textsuperscript{\normalfont 2}, Xiaodong Liu\textsuperscript{\normalfont 2}, Payal Bajaj\textsuperscript{\normalfont 2}, \\
\textbf{Yiqing Xie\textsuperscript{\normalfont 3}\footnotemark[1], Alvin Cheung\textsuperscript{\normalfont 1}, Jianfeng Gao\textsuperscript{\normalfont 2} and Xia Song\textsuperscript{\normalfont 2}} \\
\textsuperscript{\normalfont 1}UC Berkeley, \textsuperscript{\normalfont 2}Microsoft, \textsuperscript{\normalfont 3}Carnegie Mellon University \\
\textsuperscript{\normalfont 1}\small \texttt{\{gly,akcheung\}@berkeley.edu} \\
\textsuperscript{\normalfont 2}\small \texttt{\{chenyan.xiong,xiaodl,payal.bajaj,jfgao,xiaso\}@microsoft.com} \\
\textsuperscript{\normalfont 3}\small  \texttt{yiqingxi@andrew.cmu.edu}
}

\begin{document}
\maketitle

\begin{abstract}
This paper explores the effectiveness of model-generated signals in improving zero-shot generalization of text-to-text Transformers such as T5.
We study various designs to pretrain T5 using an auxiliary model to construct more challenging token replacements for the main model to denoise.
Key aspects under study include the decoding target, the location of the RTD head, and the masking pattern.
Based on these studies, we develop a new model, METRO-T0, which is pretrained using the redesigned ELECTRA-Style pretraining strategies and then 
prompt-finetuned on a mixture of NLP tasks.
METRO-T0 outperforms all similar-sized baselines on prompted NLP benchmarks, such as \textit{T0 Eval} and \textit{MMLU}, and rivals the state-of-the-art T0$_\textsc{11B}$ model with only \textbf{8\%} of its parameters. 
Our analysis on model's neural activation and parameter sensitivity reveals that the effectiveness of METRO-T0 stems from more balanced contribution of parameters and better utilization of their capacity.
The code and model checkpoints are available at \url{https://github.com/gonglinyuan/metro_t0}.
\end{abstract}
\section{Introduction}\label{sec:introduction}

% An important objective of artificial intelligence is to create models that can generalize to unseen tasks. 
% \cx{Metro-T0 or METRO-T0?}
% \linyuan{We currently use Metro-T0 because METRO-T0 looks too long in this font. Shall we switch to METRO-T0?} \cx{yes}
Recent work in NLP has shown that pretrained language models have made noteworthy progress toward generalization to unseen tasks. Despite being pretrained on only language modeling objectives, large language models can perform reasonable zero-shot generalization given natural language instructions, i.e. prompts~\citep{radford2019language,brown2020language}. Further research shows that finetuning language models on a mixture of tasks with prompt templates enhances their performance on held-out new tasks~\citep{sanh2022multitask,wei2021finetuned}.
In recent years, two significant research paths have emerged in the field of pretrained language models: one seeks to improve generalization either by scaling up the model, increasing parameters, data, and compute, or by refining prompts. Another divergent yet complementary approach focuses on augmenting the efficiency of pretraining, particularly in the context of BERT-style models. This approach has been proven to significantly improve pretraining efficiency through the use of model-generated pretraining signals, as evidenced by ELECTRA~\citep{clark2020electra}, COCO-LM~\citep{meng2021coco}, and METRO-LM~\citep{bajaj2022metro}. However, this improvement has primarily been witnessed in single-task supervised finetuning settings. Our work seeks to bridge these two areas of research. We present a novel method that enhances the pretraining efficiency of T5, a widely used encoder-decoder Transformer in prompt-based learning, by utilizing ELECTRA-Style model-generated signals.

% \linyuan{AC mentioned ``large-scale'' in the meta review, implying that a ``large-scale'' benchmark is important to show the significance of our contributions. I would prefer to keep ``large-scale''. Feel free to switch back to your version.}

% In this paper, we explore the use of efficient pretraining methods in training text-to-text Transformers and evaluate the pretrained model on large-scale prompting benchmarks. Specifically, we pretrain a T5-style Transformer encoder-decoder model using signals generated by an auxiliary masked language model.
%v2
% \cx{we need to lead with the challenge first and then solving it.} \linyuan{Added a sentence:}
Our preliminary studies, however, encountered many challenges in pretraining T5 with model-generated signals, particularly in designing an effective objective to train the decoder and ensuring training stability.
To address these challenges, we study the impact of key components in this pretraining scheme,  
such as the decoding target, the location of the Replace Token Detection (RTD) task, and the masking pattern. Then we redesign the pretraining algorithm to solve training stability issues, thus bringing in the 
benefits of ELECTRA-style pretraining to T5-style Transformer encoder-decoder models.
The pretrained model is then finetuned
on a family of multi-task training mixtures of NL-prompted dataset, which has previously been used to train the T0 models~\citep{sanh2022multitask}. Our model, METRO-T0, is a \textit{T0} model pretrained with \underline{M}odel generated d\underline{E}noising \underline{TR}aining \underline{O}bjective. 
% , is then evaluated on large-scale NL-prompted benchmarks, including \textit{T0 Eval} and \textit{MMLU}~\cite{hendrycks2020measuring}.
% \cx{this paragraph is a little too verbose and too technically detailed.} \linyuan{Removed the name of ``METRO-T5'' and the name of the prompt-finetuning dataset. However, I think we need to mention the name ``T0'' at least once in this paragraph, so establish this important relationship with prior work}

\begin{figure*}[t]
\centering
\includegraphics[width=0.98\textwidth]{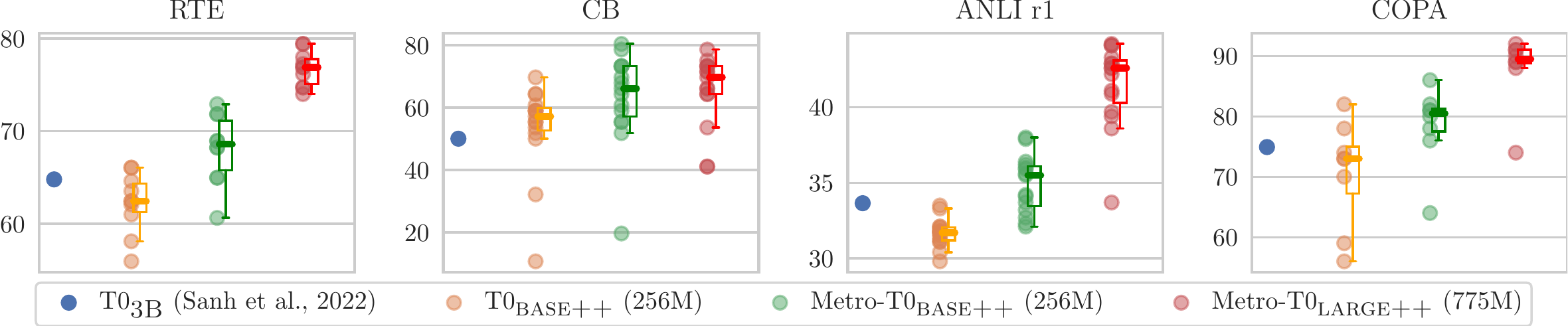}
\caption{Prompt learning results of METRO-T0 versus our T0 baseline and T0$_\textsc{3B}$ by \citet{sanh2022multitask} on 4 tasks in the \textit{T0 Eval} benchmark. Each point denotes the accuracy using one prompt template, except that the median accuracy over all templates of T0$_\textsc{3B}$ is indicated by the \textcolor{tab10blue}{\textbf{blue}} point.
% With only \textbf{256M} parameters, METRO-T0$_\textsc{base++}$ outperforms both our T0 baseline and T0$_\textsc{3B}$ with \textbf{3B} parameters. Moreover, METRO-T0$_\textsc{large++}$ has even better performance with 775M parameters.
The plots of other tasks are in \Cref{sec:appendix_full_t0}.}
\label{fig:intro}
\end{figure*}

Experimental results show that METRO-T0 is highly \textit{parameter efficient}. It consistently outperforms similar-sized baselines on all NL-prompted benchmark we evaluated upon. As shown in \Cref{fig:intro}, METRO-T0$_\textsc{Base++}$ outperforms T0$_\textsc{3B}$~\citep{sanh2022multitask} with only {7\%} of its parameters on the \textit{T0 Eval} benchmark. Moreover, METRO-T0++$_\textsc{Large++}$ rivals {14x} larger T0++$_\textsc{11B}$, the state-of-the-art in prompt-based learning.
 Our method is also \textit{compute efficient}: METRO-T0 pretrained for 500k steps has similar performance as its T0 counterpart pretrained for 2M steps. 

% \cx{we can start by saying that we also provide in-depth analysis of ELECTRA-style pretraining's impact on models that provide more insights in why it works. something more general and highlight this is also new contribution to our knowledge of pretraining.} \linyuan{Added.}
% Moreover, our research demystifies the empirical success of METRO-based pretraining, casting new light on an area previous studies only skimmed. 
To further understand the benefit of METRO pretraining, we conduct two studies on the pretrained METRO-T0 model, analyzing its neural activation and parameter sensitivity.
The studies show that model-generated signals balance the contribution of each NN parameter and reduce the number of under-activated neurons by 55\%, indicating that a key source of the improved pretraining efficiency is better utilization of network parameters.
% \cx{you can rewrite the last sentence of Abstract by making this past sentence more concise.}

% \linyuan{We planned to add a list of key contributions of this paper if we have space. Shall we add it?} \cx{no i hate it.}
\section{Related Work}\label{sec:related_work}

% Our work operates at the intersection of many broad areas of research, including big language models~\citep{devlin2019bert,lewis2019bart,raffel2019t5,brown2020language}, prompting~\citep{sanh2022multitask,wei2021finetuned}, and compute-efficient pretraining~\citep{bajaj2022metro,clark2020electra}. In this section, we discuss how our work relates to the relevant studies in each area.

\paragraph{Prompt-based  learning with language models.} \textit{Prompt-based learning} allow language models to handle a wide range of tasks with no training data (zero-shot) or a few training data (few-shot), by leveraging natural language instructions and task demonstrations as context~\citep{radford2019language,brown2020language}. \citet{raffel2019t5} proves the effectiveness of prompt-based learning as a framework of multi-task learning for text-to-text Transformers such as T5.
% Prompts can either consist of manually selected discrete tokens (hard prompts)~\citep{gao2021making}, or special tokens with continuous embeddings (soft prompts)~\citep{hacohen2019power}.
% \linyuan{Can we just omit mentions of ``soft prompts''? It is not mentioned by T0 or FLAN, seems less relevant to us.} \cx{We should mention it very briefly, cite those papers, and then briefly state our focus is on hard prompt and why. This is actually a main goal of related work, to pinpoint where this work is in the current research space.}
LMs are usually finetuned with NL instructions to improve their performance and usability. Such a procedure is called prompt-finetuning. The finetuning data comes from aggregated mixtures of NLP tasks~\citep{sanh2022multitask,wei2021finetuned}, dialogs~\citep{chung2210scaling}, or even chain-of-thoughts~\citep{wei2022chain}.
% Our study focuses on the simplest yet effective setup: hard prompt with prompt-finetuning on the T0~\citep{sanh2022multitask} task mixture.
Our work aims to improve the zero-shot generalization of T5-like text-to-text LMs in prompt-based learning by efficient and effective pretraining strategies.
% a prompting setup, but our primary technical focus is on the pretraining algorithm instead of prompts.

\paragraph{Efficient pretraining using model-generated signals.} Training big language models require substantial computational resources. This paper is part of a line of research that improves the pretraining efficiency of LMs using model-generated signals, i.e., METRO~\cite{bajaj2022metro}, pioneered by ELECTRA~\citep{clark2020electra}, a Transformer encoder pretrained using signals generated by an auxiliary BERT. Various studies~\citep{meng2021coco,meng2022pretraining,chi2021xlm,fang2022corrupted} show that an auxiliary model can generate informative training signals that greatly improve the efficiency and effectiveness of BERT-like Transformer \textit{encoder} models, as evaluated on supervised single-task benchmarks like GLUE~\citep{wang2018glue}. Compared with these works, we use model-generated signals to pretrain T5-like Transformer \textit{encoder-decoder} models and evaluate this model on large-scale NL-prompted benchmarks.

% \linyuan{Maybe another session for analysis? Sensitivity, neurons, etc.} \cx{this decision can wait till you have other parts of this paper ready.}

% \linyuan{Any missing citations?}

% \linyuan{Do we need a paragraph for our analysis methods? Neuron activations, and sensitivity. Seems that we are running out of space.}
\section{Preliminaries}\label{sec:method}

This section provides an overview of T5 and METRO-style pretraining.
% , and introduces our pretraining method.
% Then we explain how the pretrained model is prompt-finetuned to create METRO-T0/T0+/T0++.

\subsection{Text-to-Text Transformers}

Our models are based on the T5 framework~\citep{raffel2019t5}. T5 is a text-to-text Transformer pretrained on natural language corpus.
% This subsection is a brief introduction to the pretraining and finetuning pipelines of T5.
% and the achitectural upgrades we made.

\paragraph{T5 Pretraining.} T5 is a Transformer encoder-decoder language model pretrained by modeling corrupted spans of subword tokens. The noisy input is constructed by replacing consecutive spans of tokens in the input by distinct ``sentinel'' tokens, e.g., 
$X^\text{noise}=[x_1^\text{orig},...,\texttt{[M]}^{i:j},...x_n^\text{orig}]$, where the sentinel token is denoted by $\texttt{[M]}^{i:j}$. Then the pretraining task is to generate the deleted tokens using the Transformer decoder, conditioned on $X^\text{noise}$ as input to the Transformer encoder:
% The training uses standard teacher-forcing with the decoder provided with correct tokens in all previous positions.

\begin{equation}
\begin{aligned}[]
    [x_1^\text{orig},\dots \texttt{[M]}^{i:j},\dots x_n^\text{orig}]  \xrightarrow{\text{Encoder}} \boldsymbol{H}^\text{enc} \\ \boldsymbol{H}^\text{enc} \xrightarrow{\text{Decoder}} [\texttt{[M]}^{i:j},x_i^\text{orig},...,x_j^\text{orig}]. 
    \end{aligned}\label{eq:t5}
\end{equation}

\paragraph{Text-to-Text Formulation of Downstream Tasks.} T5 supports multitask learning on a diverse set of downstream tasks---including classification, question answering, and summarization---by casting all these tasks into a \textit{text-to-text} format, where the encoder is fed with the text input and the decoder is then asked to generate the target prediction.

\paragraph{Text-to-Text Prompt-Finetuning.} A pretrained text-to-text Transformer can then be finetuned to enhances its performance on held-out new tasks. The finetuning corpus is usually a multi-task mixture of NLP datasets, where each input-output pair is an example formatted with an NL prompt template. The finetuning procedure is standard seq2seq learning: the input sequence is fed to the encoder, and the target sequence serves as the ground truth to compute the cross-entropy loss of the decoder output.

\subsection{Model-Generated Pretraining Signals}

In this subsection, we discuss techniques involving model-generated pretraining signals in prior work.

\paragraph{Replace token detection (RTD)} is the training objective used to train ELECTRA~\citep{clark2020electra}. The RTD input is a noisy text sequence $X^\text{noise}$, generated by an auxiliary masked language model (MLM) like BERT. The token $x_i^\text{noise}$ in each masked position of the text sequence is sampled from the predicted probability of the auxiliary model $p_\text{MLM}(x_i^\text{noise}|\mathbf{h}_i^\text{aux})$, while the token in each unmasked position $x_j^\text{noise}$ is copied from the original text $x_j^\text{orig}$. The main model, a Transformer encoder, is pretrained to denoise the noisy input by classifying whether each token is replaced by the auxiliary model or from the original text.

{\fontsize{10.2}{11.2}\selectfont\begin{align}
&X^\text{orig} \xrightarrow{\text{Random Mask}} [x_1^\text{orig},\dots \texttt{[M]},\dots x_n^\text{orig}]; \\
&[x_1^\text{orig},\dots \texttt{[M]},\dots x_n^\text{orig}] \xrightarrow{\text{Auxiliary}} X^\text{noise}; \\
&X^\text{noise}  \xrightarrow{\text{Model}} \boldsymbol{H}  \xrightarrow{\text{RTD Head}} \mathbbm{1}(x_i^\text{orig}=x_i^\text{noise}).
\end{align}}

Prior work show that the RTD objective is more efficient than the MLM objective, resulting in significant performance improvement for pretrained Transformer encoders~\citep{clark2020electra}. However, replacing MLM with RTD turns the generative model into a discriminative model, hindering the model's ability to perform generation.

\paragraph{Corrective language modeling (CLM)} restores the generation capability of a Transformer encoder model pretrained with RTD~\citep{meng2021coco}. The CLM objective is trained alongside the RTD objective in a multi-task manner, so the CLM input is the same as the RTD input $X^\text{noise}$. The model is pretrained to recover the original text $X^\text{orig}$.

\begin{align}
X^\text{noise}  \xrightarrow{\text{Model}} \boldsymbol{H} \xrightarrow{\text{CLM Head}} X^\text{orig}.
\end{align}

\section{Method}\label{sec:method_metro}
\begin{figure*}[ht]
\centering
\includegraphics[width=0.98\textwidth]{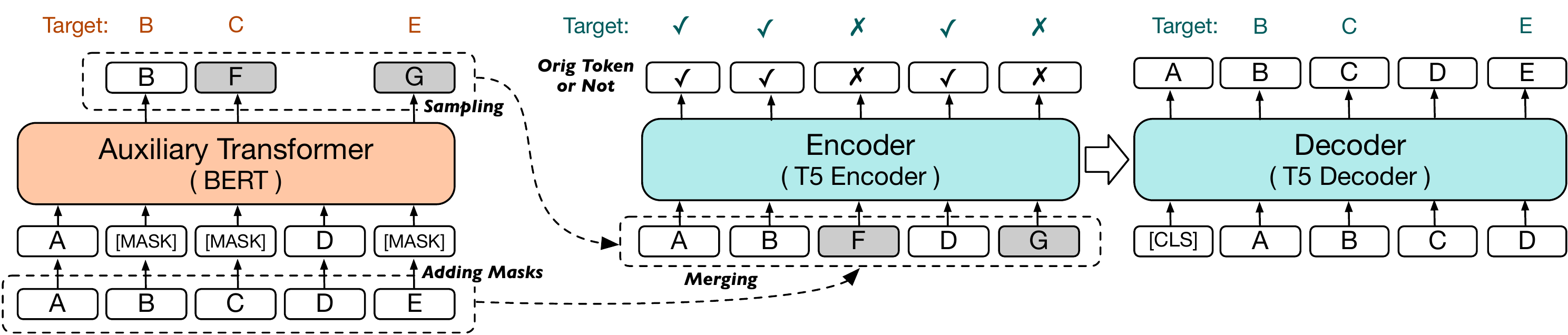}
\caption{The architecture of METRO-T0 during pretraining using BERT as the auxiliary model to generate signals.}
\label{fig:model}
\end{figure*}

In this section, we present the algorithm to train our model, METRO-T0.

\subsection{Pretraining Objective Design}\label{sec:method_objectives}

METRO-T0 is jointly pretrained with two objectives: the RTD objective, enhancing performance through model-generated signals, and the CLM objective, enabling text-to-text generation akin to T5.
The pretraining algorithm is
illustrated in \Cref{fig:model}. METRO-T0 uses a BERT-style MLM encoder as the auxiliary model and a T5-style encoder-decoder as the main model. The overall pretraining procedure is:

{\fontsize{10.2}{11.2}\selectfont\begin{align}
    &X^\text{orig} \xrightarrow{\text{i.i.d. Random Mask}} [x_1^\text{orig},\dots \texttt{[M]},\dots x_n^\text{orig}]; \\
&[x_1^\text{orig},\dots \texttt{[M]},\dots x_n^\text{orig}] \xrightarrow{\text{Auxiliary}} X^\text{noise}; \label{eq.random_mask}\\    
    & X^\text{noise}  \xrightarrow{\text{Encoder}} \boldsymbol{H}^\text{enc}  \xrightarrow{\text{RTD Head}}  \mathbbm{1}(x_i^\text{orig}=x_i^\text{noise});   \label{eq.rtd} \\    
    &  \boldsymbol{H}^\text{enc} \xrightarrow{\text{Decoder}} \boldsymbol{H}^\text{dec}   \xrightarrow{\text{CLM Head}}  X^\text{orig}.  \label{eq.clm}
\end{align}}

The auxiliary model receives inputs constructed by randomly masking tokens in the original text $X^\text{orig}$, and makes MLM predictions, which are used to create noisy inputs $X^\text{noise}$ for the main model. The main model is pretrained using two objectives: \textbf{(a)} the RTD objective on the encoder outputs $\boldsymbol{H}^\text{enc}$, which aims to identify whether each token was replaced by the auxiliary model or not, and \textbf{(b)} the CLM objective, which aims to recover the original text $X^\text{orig}$ through the decoder. During pretraining, the weighted average of three losses is optimized:

{\fontsize{10.2}{11.2}\selectfont\begin{align}
\mathcal{L}_\text{MLM} &= -\mathbb{E}_{i\in \mathcal{M}}\log p_\text{MLM}(x_i^\text{orig}|\mathbf{h}_i^\text{aux}), \\
\mathcal{L}_\text{RTD} &= -\mathbb{E} \log p_\text{RTD} (\mathbbm{1}(x_i^\text{orig} = x_i^\text{noise} ) | \mathbf{h}_i^\text{enc}), \\
\mathcal{L}_\text{CLM} &= -\mathbb{E}_{i\in\mathcal{M}}\log p_\text{LM}(x_i^\text{orig}|\mathbf{h}_i^\text{dec}), \\
\mathcal{L} &= \mathcal{L}_\text{MLM} + \lambda_\text{RTD} \mathcal{L}_\text{RTD} + \lambda_\text{CLM} \mathcal{L}_\text{CLM}.
\end{align}}

\begin{table*}[t]

\centering
\small 

\begin{tabular}{lll}
\toprule
\textbf{Original Sentence} & & Thank you for inviting me to your party last week \\ \midrule
\textbf{Auxiliary Model} & \textit{Input}& Thank you \texttt{[M]} \texttt{[M]} me to your party \texttt{[M]} week \\
& \textit{Output} & \textcolor{white}{Thank you} for giving \textcolor{white}{me to your party} apple \textcolor{white}{week} \\ \midrule
\textbf{Main Model} & \textit{Input} & Thank you for giving me to your party apple week \\ \midrule
\textbf{Decoding}  & \textit{Masked Tokens Only} & for inviting last \\
\textbf{Target}& \textit{All Tokens} & Thank you for inviting me to your party last week \\ 
& \textit{All Tokens, Masked Loss}$\bigstar$ & \textcolor{lightgray}{Thank you} for inviting \textcolor{lightgray}{me to your party} last \textcolor{lightgray}{week} \\ \bottomrule
\end{tabular}

\caption{
Examples of encoder inputs and decoder targets of different ways to configure the denoising task. \texttt{[M]} denotes a shared mask token. The auxiliary MLM model predicts one token for each \texttt{[M]}. \textcolor{lightgray}{Grayed-out} tokens are part of the target fed into the decoder but not included in pretraining loss.
}
\label{tab:demo}
\end{table*}

%v1
% Then we compare various alternatives of METRO-T5 and discuss our design choices and why they are different from prior works.
%v2
% However, this method is not the only way to pretrain a text-to-text Transformer using RTD and CLM.
%v3
In crafting METRO-T0's pretraining algorithm, we explored various alternatives before finalizing our design.
For example, an alternative method could train RTD objectives on \textit{decoder} outputs or use a masking pattern other than i.i.d. random sampling. In the rest of this section, we will explain our design choices and the reasons behind them.

% \linyuan{split into 2 paragraphs}

\paragraph{Decoding Target.} \Cref{tab:demo} shows three variants of decoding targets: \textit{``masked tokens only''}, \textit{``all tokens''}, and \textit{``all tokens masked loss''}.
% v1
% The first variant, where the decoding target consists of \textit{masked tokens only}, is the one most similar to T5. However, this target makes pretraining impossible because it is ill-formed: the decoder has no information to distinguish between the unmasked tokens (e.g., ``\texttt{you}'') or those correctly predicted by the auxiliary model in the masked positions (e.g. ``\texttt{for}''), so it is impossible for the decoder to know whether each token should be decoded or not.
% v2
% The T5-style \textit{``masked tokens only''} target makes pretraining impossible because it is ill-formed: the decoder has no information to distinguish between the unmasked tokens (e.g., ``\texttt{you}'') and those correctly predicted by the auxiliary model in the masked positions (e.g. ``\texttt{for}'') (See \Cref{sec:appendix_ill_form}).

% v3
Pretraining with the T5-style \textit{``masked tokens only''} target proves unfeasible due to its ill-formed nature. The decoder cannot distinguish between unmasked tokens (e.g., \textit{``you''}) and those correctly predicted by the auxiliary model in masked positions (e.g., \textit{``for''}). Consequently, a single source sequence may correspond to multiple correct target sequences, introducing ambiguity and impeding effective pretraining. A detailed example is provided in \Cref{sec:appendix_ill_form}.
% The second variant and the third variant avoid this issue because the decoding target is \textit{all the tokens} in the original sentence. METRO-T5 adopts the \textit{``masked loss''} style of target, where the cross entropy loss is computed and averaged on masked tokens only, so that the model is not trained on unmasked tokens which are trivial copy-and-pastes.

The \textit{``all tokens''} target is inefficient, as the cross entropy loss is averaged on all tokens, including unmasked tokens where the tasks are trivial copy-and-pastes.
Therefore, METRO-T0 uses \textit{``all tokens masked loss''}, where the loss is averaged on masked tokens only.

\paragraph{Location of the RTD Head.}
% v1
% The Transformer encoder-decoder architecture of T5 has two Transformer stacks: the encoder and the decoder.
% v2
% It is an important choice whether the RTD head should be applied to the outputs of the Transformer \textit{encoder} or the \textit{decoder}.
% v3
We consider two choices to place the RTD head: on the outputs of the Transformer \textit{encoder} or \textit{decoder}.
\textit{Decoder RTD} at position $i$ requires the information of the $i$-th token of the encoder input, but this information is absent from the input of the decoder. Consequently, the decoder needs a long attention path to connect position $i$ of the encoder. This complexity defeats the purpose of RTD in providing a simpler task to stabilize optimization, making pretraining unstable in practice~\citep{zhenhui2020mcbert}. Therefore, METRO-T0 uses \textit{encoder RTD}.

\paragraph{Masking Pattern on Auxiliary.} 
% \citet{raffel2019t5} studies the performance of span-corruption objective with different average masked span lengths and shows that longer span lengths (e.g., 5 tokens) yield better performance in natural language generation.
When can use either T5-style \textit{contiguous span masking} or BERT-style \textit{i.i.d. random masking} to generate the MLM input for the auxiliary model.
However, using \textit{contiguous span masking} in METRO-T0 pretraining leads to label leakage. At position $i$ during teacher-forced training, the decoder has access to the ground truth $X^\text{orig}$ before position $i$. It can compare $x^\text{orig}_{i-1}$ with $x^\text{noise}_{i-1}$. If the two disagree, it is likely the following position $i$ is also masked out. As a result, the model can exploit this shortcut to achieve high RTD accuracy without learning meaningful representations of natural languages. Therefore, METRO-T0 uses \textit{i.i.d. random masking}.

% \paragraph{Architectural upgrades over T5.}

\subsection{Architectural Upgrades over T5}\label{sec:method_upgrades_t5}

% \linyuan{add more details}

We incorporate model architecture changes that have been proved to be beneficial in earlier works.

The vanilla T5 exclusively uses relative positional embeddings, while the vanilla BERT~\citep{devlin2019bert} model relies solely on absolute positional embeddings. However, recent research by \citet{luo2022your} suggests that using only relative positional embeddings may not yield optimal results. Consequently, in line with the practices in COCO-LM~\citep{meng2021coco} and METRO-LM~\cite{bajaj2022metro}, we use absolute positional embeddings in addition to relative position embeddings in our model.

We also introduce a change in how layer normalization is combined with residual connections. Rather than using T5's Pre-LayerNorm approach (defined as $x \mapsto x + f(\mathrm{LN}(x))$ where $f$ is either multi-head attention or MLP), our model adopts a Post-LayerNorm design ($x \mapsto \mathrm{LN}(x + f(x))$). The Post-LayerNorm vs. Pre-LayerNorm debate is ongoing in the field, but we use Post-LayerNorm, which typically resulted in better performance on downstream tasks in our studies.

% \paragraph{METRO-T0.}

\subsection{Prompt-Finetuning}\label{sec:method_prompt_tuning}

The model pretrained using the method described above is called METRO-T5. After pretraining METRO-T5 on an NL corpus, we discard the auxiliary model and retain the main model, which is a standard text-to-text Transformer. We finetune this model on multi-task training mixtures of NL-prompted datasets, \textit{T0/T0+/T0++ Train}~\citep{sanh2022multitask}, to obtain METRO-T0/T0+/T0++, a text-to-text Transformer that supports zero-shot generalization to held-out tasks.

\section{Experimental Setup}\label{sec:experimental_setup}

\paragraph{Model Architecture.}
Each of our models has an architecture similar to T5~\citep{raffel2019t5}. We train models in three standard setups: \textit{base}, \textit{base++}, and \textit{large++}. Our \textit{base}/\textit{base++} model has an architecture similar to T5$_\textsc{Base}$. Our \textit{large++} model has an architecture similar to T5$_\textsc{Large}$ except for some differences mentioned in \Cref{sec:method_metro}.
% \cx{need to also refer to last section to note the differences are there.} \linyuan{added.}
The auxiliary model for generating training signals is a Transformer encoder of the same hidden size as the main model but is shallower: it consists of 4 layers in \textit{base}/\textit{base++} and 6 layers in \textit{large++}. We follow \citet{clark2020electra} and share token embeddings between the main and the auxiliary model.

\paragraph{Pretraining.}
Our \textit{base} model is pretrained on English Wikipedia and BookCorpus (16GB of texts) for 131 billion tokens (512 tokens per sequence, 2,048 sequences per batch, and 125k steps).
\textit{Base++}/\textit{Large++} is the training configuration first used in RoBERTa~\citep{liu2019roberta}: pretraining on a mixed corpus of 160GB texts for a maximum 2.1 trillion tokens (512 tokens per sequence, 2,048 sequences per batch, and at most 2M steps).
% See \Cref{sec:appendix_pretrain_data} and \Cref{sec:appendix_pretrain_hparams} for more details.

\paragraph{Prompt-Finetuning.}
We finetune each of our pretrained METRO-T5 models on three multi-task mixtures: \textit{T0/T0+/T0++ Train}, using the same prompt templates and shuffling strategy as \citet{sanh2022multitask} does.  Each model is finetuned for 125k steps, using the same hyperparameters as pretraining, except the peak learning rate is reduced to 0.1x. We do not perform any checkpoint selection and simply use the last checkpoint at 125k steps for evaluation.
% \Cref{sec:appendix_finetune_data} and \Cref{sec:appendix_finetune_hparams} for more details.

\paragraph{Evaluation.}
We evaluate zero-shot generalization on the \textit{T0 Eval} benchmark~\citep{sanh2022multitask} and the \textit{Massive Multi-task Language Understanding} (\textit{MMLU}) benchmark~\citep{hendrycks2020measuring}. \textit{T0 Eval} consists of 11 datasets in natural language inference, coreference, word sense disambiguation, and sentence completion. \textit{MMLU} includes exam questions from 57 tasks such as maths, history, law, and medicine. For each dataset, we report accuracy on the validation split. Following GPT-3~\citep{brown2020language} and T0~\citep{sanh2022multitask}, we use rank classification for inference.

For \textit{T0 Eval}, we use the same prompt templates as T0. For MMLU, we use prompt templates from the \textit{AI2 Reasoning Challenge} (\textit{AI2-ARC})~\citep{clark2018think}, concatenated with 5 passages retrieved using T5-ANCE~\cite{anonymous2023augmenting,ni2021sentencet5} (See \Cref{sec:appendix_evaluate_mmlu} for details).
When there are multiple prompts for a dataset, we do not perform prompt selection based on the validation split, because such prompt selection will break the ``zero-shot'' evaluation. Instead, we report the average accuracy across all prompts for this dataset, following the standard practices of \citet{sanh2022multitask}.

\paragraph{Baselines.}

For a fair comparison, the main baseline is our own T0 runs. Except for METRO-style pretraining, our T0 baselines use the same Transformer architecture, pretraining data, and prompt-finetuning data, pretrained in the same computational environment.

\begin{table*}[t]
\centering
\small
\resizebox{\textwidth}{!}{
\begin{tabular}{lr*{12}{r}}
\toprule
\multirow{2}{*}{\textbf{Model}} &
\multirow{2}{*}{\textbf{Params}} &
\multicolumn{3}{c}{\textbf{NLI}} &
\multicolumn{2}{c}{\textbf{Coref.}} &
\multicolumn{3}{c}{\textbf{Compl.}} &
\multicolumn{1}{c}{\textbf{WSD}}
\\
% \hline
% \cline{3-13}
\cmidrule(lr){3-5} \cmidrule(lr){6-7} \cmidrule(lr){8-10} \cmidrule(lr){11-11}
& &
\textbf{RTE} & \textbf{CB} & \textbf{ANLI r1/r2/r3} &
\textbf{WSC} & \textbf{Wino.} &
\textbf{COPA} & \textbf{SC.} & \textbf{HS.}  &
\textbf{WiC} &
\textbf{AVG} \\
\midrule

\multicolumn{12}{l}{\textbf{Pretraining only}  }  \\ 
\midrule

% GPT-3$_{\textsc{SMALL}}$~\citep{brown2020language} & 125M & 47.70 & 0.00 & 33.40/33.20/33.60 & 59.60 & 52.00 & 66.00 & 63.30 & 33.70 & 0.00 & 38.41\\
% GPT-3$_{\textsc{MED}}$~\citep{brown2020language} & 350M & 49.80 & 32.10 & 34.20/31.90/34.00 & 56.70 & 52.10 & 68.00 & 68.50 & 43.60 & 0.00 & 42.81 \\
GPT-3$_{\textsc{13B}}$~\citep{brown2020language} & 13B & 62.80 & 19.60 & 33.20/33.50/34.40 & 64.40 & 67.90 & 84.00 & 79.50 & 70.90 & 0.00 & 50.02 \\
GPT-3$_{\textsc{175B}}$~\citep{brown2020language} & 175B & 63.50 & 46.40 & 34.60/35.40/34.50 & 65.40 & 70.20 & 91.00 & 83.20 & 78.90 & 0.00 & 54.83 \\ 
T5+LM~\citep{lester-etal-2021-power} & 11B & 53.03 & 34.34 & 32.89/33.76/33.82 & 54.09 & 50.65 & 54.88 & 27.00 & 48.16 & 50.30 & 42.99 \\

\midrule
\multicolumn{12}{l}{\textbf{Prompt Finetune on \textit{T0 Train}}  }  \\ 
\midrule
% \multicolumn{12}{l}{\hspace{0.7em}\textbf{Base Setting}: BERT Base Size, Wikipedia + Book Corpus (16GB)} \\
T0$_{\textsc{BASE}}$ & 226M & 62.85 & 45.30 & 30.82/32.37/32.14 & \textbf{62.16} & 50.77 & 70.63 & \textbf{81.03} & 24.86 & \textbf{50.78} & 49.43 \\
METRO-T0$_{\textsc{BASE}}$ & 226M & \textbf{65.18} & \textbf{45.60} & \textbf{31.64}/\textbf{32.98}/\textbf{33.81} & 55.77 & \textbf{51.07} & \textbf{70.81} & 80.97 & \textbf{25.28} & 50.69 & \textbf{49.44} \\ \midrule

% \multicolumn{12}{l}{\hspace{0.7em}\textbf{Base++ Setting}: BERT Base Size, Bigger Training Data, and/or More Training Steps} \\
T0$_{\textsc{BASE++}}$ & 256M & 62.24 & 53.45 & 31.68/32.94/34.88 & \textbf{61.73} & 51.65 & 70.63 & 87.62 & 25.88 & \textbf{51.21} & 51.26 \\
METRO-T0$_{\textsc{BASE++}}$ & 256M & \textbf{68.16} & \textbf{63.21} & \textbf{34.92}/\textbf{33.81}/\textbf{36.82} & 60.48 & \textbf{52.03} & \textbf{78.50} & \textbf{89.23} & \textbf{27.68} & 50.88 & \textbf{54.15} \\
% \midrule 50.97 & 57.97 \\
\midrule

% \multicolumn{12}{l}{\hspace{0.7em}\textbf{Larger Reference Baselines}} \\
T0$_\textsc{3B}$~\citep{sanh2022multitask} & 3B & 64.55 & 45.36 & 33.84/33.11/33.33 & \textbf{65.10} & 50.97 & 72.40 & 84.03 & 27.29 & 50.69 & 50.97 \\
METRO-T0$_{\textsc{LARGE++}}$ & 775M & \textbf{76.75} & \textbf{65.48} & \textbf{41.49}/\textbf{36.29}/\textbf{40.18} & 60.58 & \textbf{54.51} & \textbf{88.00} & \textbf{94.07} & \textbf{29.31} & \textbf{50.97} & \textbf{57.97} \\
\midrule
T0$_\textsc{11B}$~\citep{sanh2022multitask} & 11B & 80.83 & 70.12 & 43.56/38.68/41.26 & 61.45 & 59.94 & 90.02 & 92.40 & 33.58 & 56.58 & 60.77 \\
\midrule
\multicolumn{12}{l}{\textbf{Prompt Finetune on \textit{T0+ Train}}  }  \\ 
\midrule
% \multicolumn{12}{l}{\hspace{0.7em}\textbf{Base Setting}: BERT Base Size, Wikipedia + Book Corpus (16GB)} \\ 
T0+$_{\textsc{BASE}}$ & 226M & 63.57 & \textbf{48.93} & 31.76/32.92/33.02 & \textbf{60.96} & \textbf{51.93} & \textbf{72.38} & \textit{81.71} & \textit{40.11} & \textbf{51.32} & 51.69 \\
METRO-T0+$_{\textsc{BASE}}$ & 226M & \textbf{70.56} & 47.08 & \textbf{33.05}/\textbf{34.53}/\textbf{34.37} & 57.98 & 51.75 & 69.13 & \textit{\textbf{83.08}} & \textit{\textbf{49.00}} & 50.78 & \textbf{52.85} \\ \midrule

% \multicolumn{12}{l}{\hspace{0.7em}\textbf{Base++ Setting}: BERT Base Size, Bigger Training Data, and/or More Training Steps} \\
T0+$_{\textsc{BASE++}}$ & 256M & 68.30 & 60.24 & 33.77/34.31/35.00 & 60.96 & 51.59 & 70.00 & \textit{89.29} & \textit{56.10} & 51.39 & 55.54 \\
METRO-T0+$_{\textsc{BASE++}}$ & 256M & \textbf{71.44} & \textbf{60.71} & \textbf{36.91}/\textbf{35.24}/\textbf{36.46} & \textbf{62.21} & \textbf{54.08} & \textbf{78.88} & \textit{\textbf{90.29}} & \textit{\textbf{67.57}} & \textbf{51.60} & \textbf{58.67} \\

\midrule
METRO-T0+$_{\textsc{LARGE++}}$ & 775M & 81.26 & 70.00 & 45.06/38.59/42.35 & 60.67 & 57.52 & 90.50 & \textit{95.41} & \textit{83.82} & 52.32 & 65.23 \\ \midrule
T0+$_\textsc{11B}$~\citep{sanh2022multitask} & 11B & 67.47 & 59.20 & 43.45/39.77/40.76 & 62.24 & 59.94 & 92.24 & \textit{96.43} & \textit{86.13} & 55.02 & 63.88 \\

\midrule
\multicolumn{12}{l}{\textbf{Prompt Finetune on \textit{T0++ Train}}  }  \\
\midrule
% \multicolumn{12}{l}{\hspace{0.7em}\textbf{Base Setting}: BERT Base Size, Wikipedia + Book Corpus (16GB)} \\
T0++$_{\textsc{BASE}}$ & 226M & 69.06 & 48.39 & 31.90/33.61/33.94 & \textit{55.72} & \textit{51.15} & \textit{\textbf{76.06}} & \textit{82.55} & \textit{39.62} & \textit{63.18} & 53.20 \\
METRO-T0++$_{\textsc{BASE}}$ & 226M & \textbf{72.04} & \textbf{58.63} & \textbf{33.85}/\textbf{35.29}/\textbf{36.57} & \textit{\textbf{56.11}} & \textit{\textbf{52.15}} & \textit{74.06} & \textit{\textbf{83.65}} & \textit{\textbf{48.66}} & \textit{\textbf{64.29}} & \textbf{55.94} \\ \midrule

% \multicolumn{12}{l}{\hspace{0.7em}\textbf{Base++ Setting}: BERT Base Size, Bigger Training Data, and/or More Training Steps} \\
T0++$_{\textsc{BASE++}}$ & 256M & \textbf{77.87} & 63.10 & 36.15/34.61/38.18 & \textit{56.44} & \textit{51.78} & \textit{75.38} & \textit{89.33} & \textit{55.95} & \textit{65.53} & 58.57 \\
METRO-T0++$_{\textsc{BASE++}}$ & 256M & 77.80 & \textbf{69.52} & \textbf{39.69}/\textbf{36.61}/\textbf{40.08} & \textit{\textbf{61.44}} & \textit{\textbf{54.55}} & \textit{\textbf{83.88}} & \textit{\textbf{90.88}} & \textit{\textbf{68.54}} & \textit{\textbf{67.59}} & \textbf{62.78} \\

\midrule
METRO-T0++$_{\textsc{LARGE++}}$ & 775M & 83.68 & 74.88 & 46.84/40.37/44.95 & \textit{71.83} & \textit{62.75} & \textit{92.63} & \textit{95.65} & \textit{83.74} & \textit{70.49} & 69.80 \\ \midrule
T0++$_\textsc{11B}$~\citep{sanh2022multitask} & 11B & 85.31 & 75.69 & 47.07/42.18/44.09 & \textit{70.29} & \textit{66.42} & \textit{93.71} & \textit{96.49} & \textit{86.11} & \textit{70.02} & 70.67 \\

 \bottomrule
\end{tabular}
}
\caption{Prompt learning results on the \textit{T0 Eval} dataset. ``Wino.'', ``SC.'', and ``HS'' refer to Winogrande, StoryCloze, and HellaSwag, respectively. All reported datasets use accuracy as their metric. \textit{Italic} results are produced under the supervised setting. Others are under the zero-shot setting. Each row without a citation contains experimental results from models trained by us (our T0 baseline and METRO-T0), while each row with a citation contains experimental results from the cited paper (GPT-3, Google T5, and the original T0).}
\label{tab:t0}
\end{table*}

\begin{table}[t]
\centering
\small 
\begin{tabular}{lrr}
\toprule
\textbf{Model} & \textbf{Params} & \textbf{MMLU} \\
\midrule
T0++$_\textsc{base}$ & 226M & 37.5 \\
METRO-T0++$_\textsc{base}$ & 226M & \textbf{38.3} \\
\midrule
Flan-T5$_\textsc{base}$~\citep{wei2022chain} & 223M & 35.9 \\
T0++$_\textsc{base++}$ & 256M & 41.7 \\
METRO-T0++$_\textsc{base++}$ & 256M & \textbf{42.7} \\
\midrule
GPT-3$_\textsc{175B}$~\citep{brown2020language} & 175B & 43.9 \\
Flan-T5$_\textsc{large}$~\citep{wei2022chain} & 750M & 45.1 \\
T0++$_\textsc{11B}$~\citep{sanh2022multitask} & 11B & 35.6 \\
METRO-T0++$_\textsc{large++}$ & 775M & \textbf{48.0} \\
\bottomrule
\end{tabular}
\caption{Prompt learning results on the \textit{MMLU} dataset. All reported results use accuracy averaged over 57 subtasks as their metric.}
\label{tab:mmlu}
\end{table}

We also compare with the \textit{reported} numbers of other language models that supports zero-shot prompting, including pretraining-only models such as GPT-3~\cite{brown2020language} and T5~\cite{raffel2019t5}, as well as prompt-finetuned models such as T0~\cite{sanh2022multitask} and Flan-T5~\cite{wei2021finetuned,chung2210scaling}. T0/T0+/T0++ is pretrained on the the C4~\citep{raffel2019t5} corpus of 800GB of texts for 1 trillion tokens and then prompt-finetuned on the \textit{T0/T0+/T0++ Train} multitask mixture after LM adaptation for 100 billion tokens. Flan-T5 is also pretrained on the C4 corpus, but finetuned on a much larger dataset of prompted multi-task mixtures, dialog, and chain-of-thoughts.

\section{Evaluation Results}\label{sec:results}

This section compares the performance of METRO-T0 and baseline models on \textit{T0 Eval} and \textit{MMLU} to demonstrate the effectiveness and efficiency of our method.
% v1
% We also analyze neural activations and parameters of different models to explain the performance improvement achieved by METRO-T0.
% v2
We also explore the reason behind METRO-T0's effectiveness through detailed model analysis.

\subsection{Main Results}\label{sec:results_main_results}

\Cref{tab:t0} presents the experimental results on \textit{T0 Eval}, and \Cref{tab:mmlu} presents the experimental results on \textit{MMLU}. These results show that:

\paragraph{METRO-T0 is highly parameter efficient,} as it rivals or even outperforms much larger models in zero-shot generalization. METRO-T0$_\textsc{base++}$, having only 256M parameters, outperforms T0$_\textsc{3B}$~\citep{sanh2022multitask} with only \textbf{7\%} of its parameters.
Also, METRO-T0$_\textsc{large++}$, having only 775M parameters, outperforms T0$_\textsc{3B}$ by 7pts and is only 2.8pts behind T0$_\textsc{11B}$, a 14x larger model.

% \paragraph{METRO-T0 has state-of-the-art prompting performance.} METRO-T0$_\textsc{large++}$ outperforms GPT-3$_\textsc{175B}$,
\paragraph{METRO-T0 often outperforms GPT-3 (175B), }
a state-of-the-art Transformer decoder LM, on both \textit{T0 Eval} and \textit{MMLU}. Compared to the 11B-parameter T0/T0+/T0++ model, a family of state-of-the-art prompt-finetuned text-to-text LM, METRO-T0/T0+/T0++ in the \textit{large++} setup has competitive or sometimes superior performance.

\paragraph{The gain stems from METRO-style pretraining.} On both benchmarks, METRO-T0 models in all setups consistently outperform our fair-comparison T0 baselines of the same model size, which were pretrained using the same corpus and configurations. This fact demonstrates that the performance improvement is not due to better hyperparameters or data engineering, but a result of using METRO-style pretraining.
Further confirmation of this argument will be provided through model analysis in \Cref{sec:analysis_neuron} and \Cref{sec:analysis_sensitivity}.

\subsection{Ablation Studies}\label{sec:results_ablation_studies}

\begin{table}[t]
\centering
\small 
\begin{tabular}{lrrr}
\toprule
\textbf{Model/Finetuning Data} & \textbf{T0} & \textbf{T0+} & \textbf{T0++} \\
\midrule
% 85
METRO-T0/T0+/T0++ & 49.44 & \textbf{54.15} & \textbf{57.97}
\\
% 100
+ CLM Loss on All Position & 49.24 & 51.05 & 53.97
\\
% 41
+ CLM with Copy Mechanism & \textbf{49.46} & 50.70 & 54.06
\\
% 80
+ RTD on Decoder & 46.75 & 48.47 & 49.20
\\
% 81
\text{ }\hspace{1em} + Projection Layer on CLM & 48.85 & 50.10 & 52.82
\\
% 99
+ Continuous Span Mask & 49.04 & 50.37 & 53.42
\\
\midrule
% 9
T0/T0+/T0++ & \textbf{49.43} & \textbf{51.69} & \textbf{53.20}
 \\
 % 11
+ All-token LM loss & 48.13 & 49.43 & 50.76
 \\
\bottomrule
\end{tabular}
\caption{Performance of METRO-T0 variations on \textit{T0 Eval}. All ablations are done in the \textit{base} pretraining setting using exactly the same prompt-finetuning pipeline.}
\label{tab:ablation}
\end{table}

In \Cref{sec:method_metro}, we discuss the choices we made to redesign the pretraining method for METRO-T0. In this subsection, we compare the empirical results of different variants of METRO-T0. \Cref{tab:ablation} shows the performance of each variant prompt-finetuned on \textit{T0/T0+/T0++ Train} and evaluated on \textit{T0 Eval}.

\paragraph{``All tokens, masked loss'' is the best decoding target.} \Cref{tab:demo} presents three possible choices for the decoding target, in which \textit{``masked tokens only''} is ill-formed and thus not suitable, as discussed in \Cref{sec:method_metro}.
\Cref{tab:ablation} compares the remaining two options and shows that computing CLM/LM loss on all positions negatively affects the downstream performance of METRO-T5/T5 by overwhelming the model with too many trivial copy-and-paste tasks. The same reasoning also applies to our decision not to use the copy mechanism~\citep{meng2021coco} in CLM heads.

\begin{figure}[t]
\centering
\begin{subfigure}[t]{0.2696551724137931\textwidth}
\centering
\includegraphics[width=\textwidth]{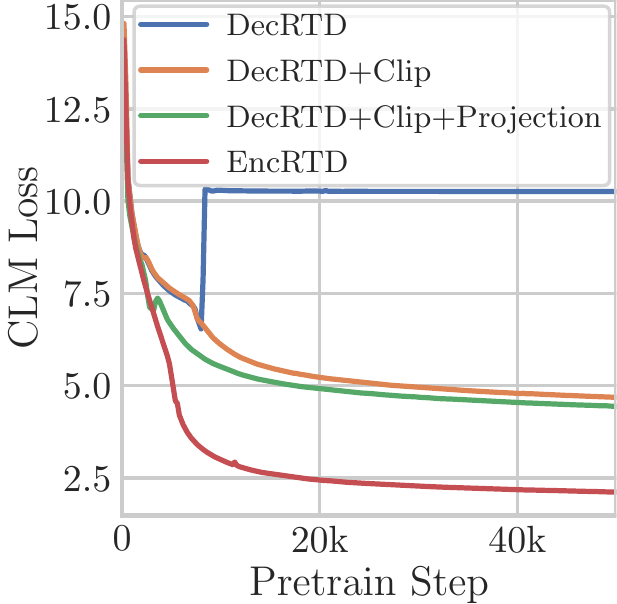}
\caption{RTD Position\label{fig:decrtd}}
\end{subfigure}
\hfill
\begin{subfigure}[t]{0.1903448275862069\textwidth}
\centering
\includegraphics[width=\textwidth]{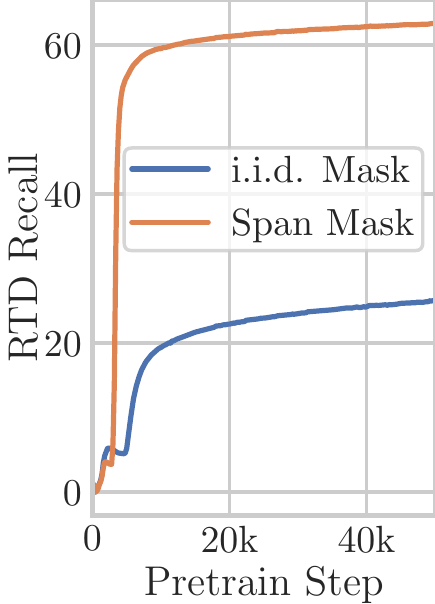}
\caption{Masking Strategy\label{fig:spanmask}}
\end{subfigure}
\caption{Pretraining behaviors of different designs.}
\label{fig:stability}
\end{figure}

\paragraph{Encoder RTD makes pretraining more stable.} \Cref{fig:decrtd} demonstrates this by comparing the loss on the CLM task during pretraining with RTD applied to the encoder (\textcolor{tab10red}{\textbf{red}} line) versus the decoder (\textcolor{tab10blue}{\textbf{blue}} line). Decoder RTD caused pretraining to diverge. While techniques such as strong gradient clipping and an additional projection layer can mitigate this issue (\textcolor{tab10orange}{\textbf{orange}} and \textcolor{tab10green}{\textbf{green}} lines), the model still has higher training loss and poorer generalization on downstream tasks as shown in \Cref{tab:ablation}.

\paragraph{Label leakage is prevented by i.i.d. masking.} \Cref{fig:spanmask} illustrates the RTD \textit{recall} (true positive rate) of METRO-T5 when using i.i.d. random masking on the auxiliary model compared to T5's continuous span masking. As discussed in \Cref{sec:method_metro}, continuous span masking leads to label leakage, resulting in easy solutions for many masked positions, as demonstrated by the more than 2x pretraining RTD recall on masked positions with \textcolor{tab10orange}{\textbf{Span Mask}}. As expected, this label leakage hurts the model's generalization ability as shown in \Cref{tab:ablation}.

% \section{Analysis}\label{sec:analysis}

% This section compares T5 pretraining and METRO-T5 pretraining regarding their compute efficiency, i.e., their downstream prompting performance given the same amount of compute. To interpret these performance/efficiency differences, we conduct analyses of neural activations and parameter sensitivity. We pretrain all T5/T0 baseline models mentioned in this section using the same setting as their METRO counterparts for fair comparisons. All models mentioned in this section are pretrained using the \textit{base++} setting.

% To merge with previous section

\subsection{Pretraining Efficiency}\label{sec:analysis_efficiency}

In this experiment, we study the pretraining efficiency of METRO-T5
% v1
% We finetune the intermediate checkpoints pretrained for 500k/1M/2M steps of T5$_\textsc{base++}$ and METRO-T5$_\textsc{base++}$ on the \textit{T0++ Train} dataset.
% v2
by comparing the intermediate checkpoints pretrained for 500k/1M/2M steps of T5$_\textsc{base++}$ and METRO-T5$_\textsc{base++}$.
We assess each checkpoint's prompt-based learning performance by finetuning on the \textit{T0++ Train} dataset and recording the average performance on \textit{T0 Eval}.

% \Cref{fig:efficiency} shows the performance of each finetuned model checkpoint on the \textit{T0 Eval} benchmark, with its pretraining wall time as its x-coordinate.
% Because we pretrain all models in the same computational environment, the pretraining wall time exactly reflects the pretraining computational cost of each model.
% Notably, each pretraining step of METRO-T5 is slower than each step of T5 due to the additional compute to train the auxiliary model and longer sequences of the decoder target.
% However, \Cref{fig:efficiency} shows that the efficiency benefits of METRO pretraining overwhelm the increased compute \textit{per step}. METRO-T0++ achieves better downstream performance at \textit{every point}. In particular, METRO-T0++ pretrained for 500k steps has a similar performance to T0++ pretrained for 2M steps, showing a \textbf{165\%} efficiency increase.
% v2
\begin{figure}[t]
\centering
\includegraphics[width=0.4144554455445545\textwidth]{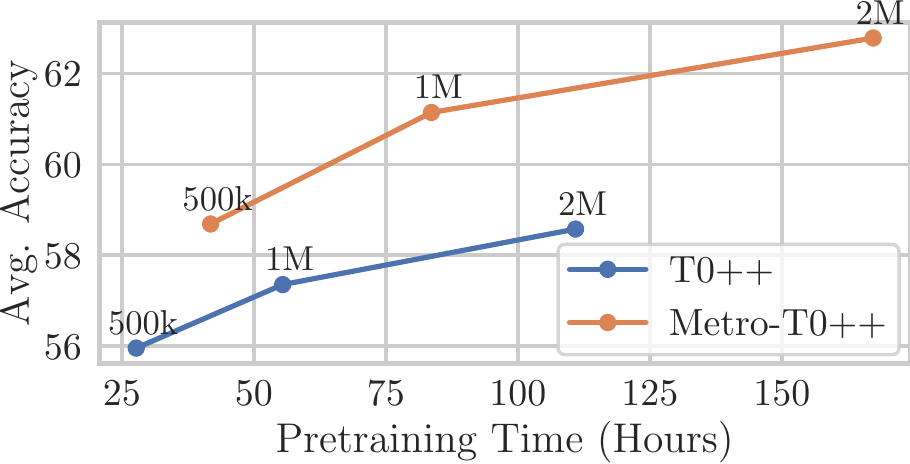}
\caption{Comparison of the pretraining efficiency of T5 and METRO-T5. Each point shows the performance of a T0++/METRO-T0++ model finetuned from a checkpoint at 500k/1M/2M pretraining steps.
% As our focus is efficiency, the x-axis shows pretraining wall time instead of training steps. The wall time exactly reflects the computational cost to pretrain each model because we pretrain all models in the same environment.
The x-axis displays the pretraining wall time, reflecting computational cost, as all models were pretrained in the identical environment.
}
\label{fig:efficiency}
\end{figure}

\Cref{fig:efficiency} shows that \textbf{METRO-T5 is more compute efficient than vanilla T5}. METRO-T0++ achieves better downstream performance at \textit{every point}. In particular, METRO-T0++ pretrained for 500k steps has a similar performance to T0++ pretrained for 2M steps, showing a \textbf{165\%} efficiency increase.

An interesting research question is: does model-generated signals simply make pretraining faster or do METRO-T5 and T5 learn different representations?

\begin{figure}[t]
\centering
\includegraphics[width=0.46\textwidth]{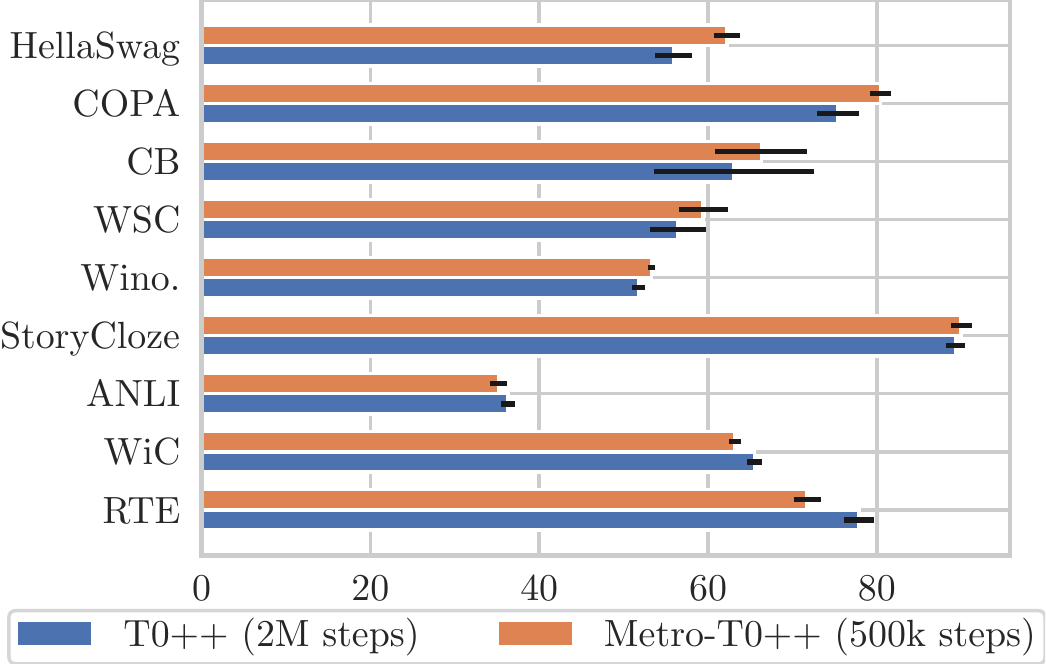}
\caption{Per-task performance of T0++ (pretrained for 2M steps) and METRO-T0++ (pretrained for only 500k steps) on \textit{T0 Eval}.
The error bars are calculated using the model's performance across prompt templates.
% The average accuracies of these two models are similar (58.57 vs. 58.68), but they are strong at different tasks.
% \cx{only describe the fig in caption. leave discussion in main contents}\linyuan{Removed. This sentence is actually already in the main text. I know having the sentence here is a ``bad'' practice that is also complained by my own advisor. I still don't understand why. When I read papers, the first thing I read are figures and tables, and informative captions that summarize the results really help.}
}
\label{fig:efficiency_compare}
\end{figure}

To answer this question, we compare the following two models by showing their performance on each task in the \textit{T0 Eval} benchmark in \Cref{fig:efficiency_compare}: \textbf{(a)} T0++ finetuned from the T5 checkpoint pretrained for 2M steps, indicated by the last \textcolor{tab10blue}{\textbf{blue}} datapoint in \Cref{fig:efficiency}; \textbf{(b)} METRO-T0++ finetuned from the METRO-T5 checkpoint pretrained for 500k steps, indicated by the first \textcolor{tab10orange}{\textbf{orange}} datapoint. Although these two models have similar average accuracies (58.57 vs. 58.68), they have different strengths, as shown in \Cref{fig:efficiency_compare}. T0++ (2M steps) outperforms METRO-T0++ (500k steps) on word-level tasks (WiC) and conventional natural language inference (ANLI and RTE), while METRO-T0++ (500k steps) has much better performance on commonsense reasoning (HellaSwag and COPA). This phenomenon implies that model-generated signals let the model learn different representations of texts, which finally result in a significant performance gap between the fully pretrained T0++ and METRO-T0++, as shown in \Cref{tab:t0}.

\subsection{Neural Activation}\label{sec:analysis_neuron}

In this subsection, and the following one, explore the extent to which the internal statistics of the neural networks quantify the differences between METRO-T5 and T5.

The first aspect we explore is neural activation. Specifically, we examine the feedforward module in each Transformer layer of METRO-T5$_\textsc{Base++}$ and T5$_\textsc{Base++}$, counting neurons that are \textit{under-activated}. A neuron is considered \textit{under-activated} if it is \textit{inactive} (exhibits zero ReLU activations) for 99.5\% of tokens within the \textit{T0++ Train} dataset. 

\begin{figure}[t]
\centering
\includegraphics[width=0.4144554455445545\textwidth]{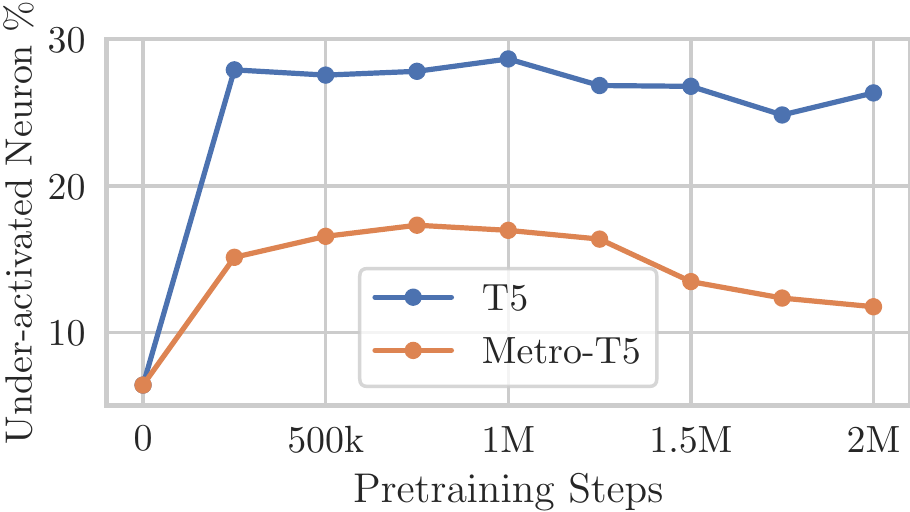}
\caption{Comparison of the percentage of under-activated neurons in T5 and METRO-T5 on \textit{T0++ train} dataset. The first point of both models (0 steps) overlap because they are the same initial model.}
\label{fig:neuron}
\end{figure}

% \caption{Comparison of neural activations and parameter sensitivity between T5 and METRO-T5.}

\Cref{fig:neuron} shows that T5 has $\sim$\textbf{2x} as many under-activated neurons as METRO-T5 at \textit{every} checkpoint. Studies suggest that such neurons can typically be pruned without substantially affecting neural network performance~\cite{channel2015polyak,pruning2016li}. So the presense of many under-activated neurons is a sign of underutilization of model capacity and computing cost.
Therefore, our findings suggest that METRO-style model-generated training signals enhance neuron utilization in METRO-T5.

\subsection{Parameter Sensitivity}\label{sec:analysis_sensitivity}

% v1
% Besides neural activation, we can also quantify the difference between T5 and METRO-T5 by comparing the sensitivity of their parameters.
% v2
% Besides transient neural activation, a more persistent determinant---model parameters---serve as a means to quantify the underlying differences between T5 and METRO-T5. Hence, we assess the sensitivity of parameters to further compare these models.
% v3
In addition to analyzing the neural activation of T5 and METRO-T5, we also examine their parameter sensitivity, which serves as another means to quantify the underlying differences between T5 and METRO-T5.

The \textit{sensitivity} of a parameter, defined in \Cref{eqn:sensitivity}, approximates the change in the loss magnitude when this parameter is completely zeroed-out. $\theta$ denotes the parameter vector and $\mathcal{L}$ denotes the loss function. $\theta_{-j}$ denotes the parameter vector $\theta$ with its $j$-th entry set to zero. The approximation is derived from the first-order Taylor expansion of $\mathcal{L}$ at $\theta$. Therefore, the sensitivity of the $j$-th parameter, denoted by $I_j$, approximates the change in the loss magnitude when this parameter is completely zeroed-out~\citep{lecun1989optimal}.
\begin{equation}
I_j = |\theta_{-j}^T\nabla_\theta \mathcal{L}(\theta)| \approx |\mathcal{L}(\theta) - \mathcal{L}(\theta - \theta_{-j})| 
\label{eqn:sensitivity}
\end{equation}

\citet{liang2022noparameter} shows that parameter sensitivity is a reliable indicator of redundancy in pretrained language models. Specifically, parameters with low sensitivity can be safely pruned with only marginal impact on the LM's downstream performance, and an LM with lower, more concentrated sensitivity is more sufficiently trained and generalizes better.

\begin{figure}[t]
\centering
\includegraphics[width=0.4446492711771177\textwidth]{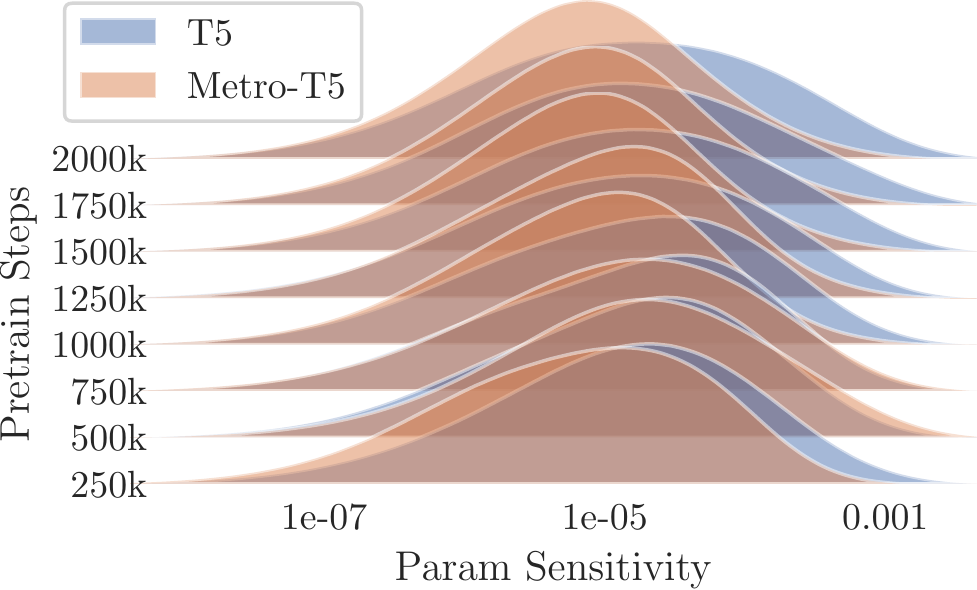}
\caption{Comparison of the parameter sensitivity distributions of T5 and METRO-T5. Each row shows the parameter sensitivity distributions of T5 and METRO-T5 at the same pretraining step, indicated by the corresponding label on the y-axis.}
\label{fig:sensitivity_2}
\end{figure}

We compare parameter sensitivity distributions of each checkpoint of METRO-T5 and T5, using gradients calculated on the \textit{T0++ Train} dataset. The result is shown in \Cref{fig:sensitivity_2}, from which we observe that the sensitivity distribution exhibits a lower variance in METRO-T5 (the \textcolor{tab10orange}{\textbf{orange}} hill in each row) than in T5 (the \textcolor{tab10blue}{\textbf{blue}} hill in each row). The difference in parameter sensitivity becomes more conspicuous when the models are trained for more steps. These observations suggest that pretraining with model-generated signals makes the sensitivity of parameters more concentrated. In other words, the amount of each parameter's contribution becomes more balanced with METRO-style pretraining, which leads to a more sufficiently trained model.

\section{Conclusion}\label{sec:conclusion}

This paper presents a new method for improving the zero-shot generalization of T5-like text-to-text Transformers by incorporating model-generated signals in the pretraining process.
% The proposed model, METRO-T0, is highly parameter efficient and compute efficient because it is more sufficiently trained using our redesigned pretraining method.
METRO-T0, the model sufficiently trained using our redesigned pretraining method, is highly parameter efficient and compute efficient.
We hope that the success of our approach could inspire further work on efficient big LM pretraining and prompt-based learning.

\section*{Limitations}
% ACL 2023 requires all submissions to have a section titled ``Limitations'', for discussing the limitations of the paper as a complement to the discussion of strengths in the main text. This section should occur after the conclusion, but before the references. It will not count towards the page limit.
% The discussion of limitations is mandatory. Papers without a limitation section will be desk-rejected without review.

% While we are open to different types of limitations, just mentioning that a set of results have been shown for English only probably does not reflect what we expect. 
% Mentioning that the method works mostly for languages with limited morphology, like English, is a much better alternative.
% In addition, limitations such as low scalability to long text, the requirement of large GPU resources, or other things that inspire crucial further investigation are welcome.
This work focuses on pretraining large language models for zero-shot generalization. Although our proposed method is more efficient than baselines, it still requires significant computational resources, specifically GPU resources. The GPU resources used and training time are detailed in \Cref{sec:appendix_implement_detail}. Our study is also limited by the computational budget, preventing us from training models as large as GPT-3 or T0$_\textsc{11B}$. However, our \textit{large++} model (775M parameters) already rivals or outperforms previous state-of-the-art models.

\section*{Ethics Statement}
This work proposes and releases language models that are pretrained on web-crawled data and finetuned on a large collection of NLP datasets. These models may perpetuate social stereotypes and disparities reflected in the training data, or accidentally reveal private information.
% Eliminating these risks from such large training corpora is infeasible at the current stage.
Mitigating these risks presents a significant open research challenge that calls for collective efforts within the NLP community.
Therefore, it is recommended to take appropriate measures to assess risks and potential harms in the application context before deployment.

\section*{Acknowledgement}

Linyuan Gong and Alvin Cheung are partially supported by the National Science Foundation through grants IIS-1955488, IIS-2027575, CCF-1723352, ARO W911NF2110339.
We thank Mingrui Shen for his support providing computing infrastructure support on the finetuning work flow, Guolin Ke for his support in establishing the data processing pipelines for our pretraining corpora, and anonymous reviewers for their constructive feedback.

\bibliography{references}
\bibliographystyle{acl_natbib}

\newpage
\onecolumn
\appendix
% \the\textwidth

% \the\parskip

% \the\abovecaptionskip

% \the\textfloatsep

% \the\dbltextfloatsep

% \the\parsep
\section{Appendix}
\subsection{Pretraining Corpus}\label{sec:appendix_pretrain_data}

Our \textit{base} model is pretrained on English Wikipedia and BookCorpus (16GB of texts). We encode the pretraining corpus with an uncased vocabulary of 32,768 BPE tokens. This setup is similar to vanilla BERT~\citep{devlin2019bert}.

Our \textit{base++/large++} model is pretrained on a mixed corpus of 160GB texts, which consists of English Wikipedia, BookCorpus, OpenWebText~\citep{Gokaslan2019OpenWeb}, CC-News~\citep{liu2019roberta}, and STORIES~\citep{trinh2018simple}. We encode the corpus with a cased vocabulary of 64,000 BPE tokens. This setup is similar to RoBERTa~\citep{liu2019roberta}, COCO-LM~\citep{meng2021coco}, and METRO-LM~\citep{bajaj2022metro}.

As a reference, T0~\citep{sanh2022multitask} models and Flan-T5~\citep{chung2210scaling} are all based on the original T5 model by \citet{raffel2019t5}. The pretraining corpus is the C4 corpus~\citep{raffel2019t5} of 800GB of texts based on CommonCrawl. They encode the corpus with a cased vocabulary of 32k BPE tokens.
\subsection{Pretraining Hyperparameters}\label{sec:appendix_pretrain_hparams}

The hyperparameters we used to pretrain METRO-T0 and our T0 baseline are listed in \Cref{tab:appendix_hparam_pretrain}.

\begin{table}[h]
\centering
\begin{tabular}{lrrr}
\toprule
\textbf{Hyperparameters} & \textbf{Base} & \textbf{Base++} & \textbf{Large++}  \\
\midrule
Encoder Layers & 12 & 12 & 24 \\
Decoder Layers & 12 & 12 & 24 \\
Auxiliary Layers$^*$ & 4 & 4 & 6 \\
Hidden Dimension & 768 & 768 & 1,024 \\
Peak Learning Rate & 4e-4 & 2e-4 & 2e-4 \\
Batch Size & 2,048 & 2,048 & 2,048 \\
Warm-Up Steps & 10,000 & 10,000 & 10,000 \\
Total Steps & 125,000 & 2,000,000 & 1,335,000 \\
Sequence Length & 512 & 512 & 512 \\
Relative Position Encoding Buckets & 32 & 32 & 64 \\
Relative Position Encoding Max Distance & 128 & 128 & 128 \\
Loss multipliers ($\lambda_\text{MLM}$, $\lambda_\text{RTD}$, $\lambda_\text{CLM}$)$^*$  & (1, 50, 1) & (1, 50, 1) & (1, 50, 1) \\
Adam $\epsilon$ & 1e-6 & 1e-6 & 1e-6 \\
Adam ($\beta_1$, $\beta_2$) & (0.9, 0.98) & (0.9, 0.98) & (0.9, 0.98) \\
Clip Norm & - & 2.0 & 2.0 \\
Dropout & 0.1 & 0.1 & 0.1  \\
Weight Decay & 0.01 & 0.01 & 0.01 \\
\bottomrule
\end{tabular}
\caption{Pretraining hyperparameters for METRO-T0 and our T0 baselines. Rows with an ``$^*$'' are specific to METRO-style pretraining and not applicable to our T0 baselines. We only train our \textit{large++} model for 1.3M steps (instead of 2M steps) due to limitations of computational budgets but it still yields impressive performance.}
\label{tab:appendix_hparam_pretrain}
\end{table}

% mask
In pretraining, we use 15\% masking ratio for the auxiliary MLM pretraining task. We create a \texttt{[MASK]} symbol for each masked token.
% The average length of masked span is \todo{XXX}.
% pretraining task weight
Each token in $X^\text{noise}$ is sampled from the softmax distribution predicted by the auxiliary model for each \texttt{[MASK]} symbol.
The weight of each pretraining objective is $\lambda_\text{MLM}=1$, $\lambda_\text{RTD}=50$, and $\lambda_\text{CLM}=1$, following \citet{meng2021coco}.
% share embedding
In both the auxiliary transformer and the main transformer, we use shared token embeddings in the embedding layer and the language modeling head.

We have three projection heads in our model: MLM head on the auxiliary transformer, RTD head on the main transformer's encoder, and CLM head on the main transformer's decoder.
Both the MLM and CLM head are a single linear transformation.
We use RoBERTa-style projection head for the RTD head, which contains a linear projection, a ReLU activation, a layer norm and another linear projection.
For the RTD on decoder (complex CLM head) ablation, we use a RoBERTa-style head as the architecture of the CLM head.
\subsection{Data for Prompt-Finetuning}\label{sec:appendix_finetune_data}

Following \citet{sanh2022multitask}, we finetune our models on three training mixtures, \textit{T0 Train} (39 datasets), \textit{T0+ Train} (49 datasets), and \textit{T0++ Train} (55 datasets), respectively. Each dataset is associated with multiple (8.03 on average) prompt templates that are used to format example instances to input and target pairs. Please refer to \citet{sanh2022multitask} for more details about our finetuning datasets.
\subsection{Prompt-Finetuning Hyperparameters}\label{sec:appendix_finetune_hparams}

Once we have METRO-T5 pretrained on a natural language corpus, we discard the auxiliary model and keep the main model, which is a standard text-to-text Transformer. We finetune this model on multi-task training mixtures of NL-prompted datasets proposed by \citet{sanh2022multitask}. Once the model parameters are initialized with pretrained METRO-T5, the finetuning procedure is standard sequence-to-sequence learning: the input sequence is fed to the encoder, and the target sequence serves as the ground truth to compute the cross-entropy loss of the decoder output. Each model is finetuned using hyperparameters listed in \Cref{tab:appendix_hparam_finetune}. Basically, we use the same hyperparameters as pretraining, except the peak learning rate is reduced to 0.1x and each target sequence is truncated to a max length of 256. We do not perform any checkpoint selection or hyperparameter selection, and simply use the last checkpoint at 125k steps of this single run for evaluation. 

\begin{table}[h]
\centering
\begin{tabular}{lrrr}
\toprule
\textbf{Hyperparameters} & \textbf{Base} & \textbf{Base++} & \textbf{Large++} \\
\midrule
Peak Learning Rate & 4e-5 & 2e-5 & 2e-5 \\
Total Steps & 125,000 & 125,000 & 125,000 \\
Source Sequence Length & 512 & 512 & 512 \\
Target Sequence Length & 256 & 256 & 256 \\
Clip Norm & - & - & - \\
\bottomrule
\end{tabular}
\caption{
Hyperparameters for prompt-finetuning METRO-T5 and our pretrained T5 baseline. All hyperparameters not mentioned in this table is the same as in the pretraining procedure.
}
\label{tab:appendix_hparam_finetune}
\end{table}

\subsection{Evaluation}

We evaluate zero-shot generalization on the \textit{T0 Eval} benchmark~\citep{sanh2022multitask} and the \textit{Massive Multi-task Language Understanding} (\textit{MMLU}) benchmark~\citep{hendrycks2020measuring}. T0 Eval consists of 11 held-out datasets in natural language inference, coreference, word sense disambiguation, and sentence completion, and details are shown in \Cref{tab:appendix_benchmark}  MMLU includes example questions from 57 tasks such as maths, history, law, and medicine. Please refer to \citet{hendrycks2020measuring} for more details about MMLU.

\begin{table}[h]
\centering
\begin{tabular}{lrll}
\toprule
& \textbf{Size} & \textbf{Task} & \textbf{Metric}  \\
\midrule
RTE & 277 & Natural language inference & Accuracy \\
CB & 56 & Natural language inference & Accuracy \\
ANLI & 3,200 & Natural language inference & Accuracy \\
WSC & 104 & Coreference resolution & Accuracy \\
Winogrande XL & 1,267 & Coreference resolution & Accuracy \\
COPA & 100 & Sentence completion & Accuracy \\
StoryCloze 2016 & 1,871 & Sentence completion & Accuracy \\
HellaSwag & 10,042 & Sentence completion & Accuracy \\
WiC & 638 & Word Sense Disambiguation & Accuracy \\

\bottomrule
\end{tabular}
\caption{The overview of the \textit{T0 Eval} benchmark for prompt learning.}
\label{tab:appendix_benchmark}
% \vspace{-0.3cm}
\end{table}

Each task in \textit{T0 Eval} or \textit{MMLU} is formulated as multiple-choice questions. We compute the log probability of each choice under the finetuned model and select the choice with the highest log probability as the prediction.
\subsection{Implementation Details}\label{sec:appendix_implement_detail}

\paragraph{Implementation} We implement our T0 baseline and METRO-T0 based on \texttt{fairseq}\footnote{\url{https://github.com/facebookresearch/fairseq}}. The prompt templates to format the finetuing data are from the \texttt{promptsource}\footnote{\url{https://github.com/bigscience-workshop/promptsource}} library~\citep{bach2022promptsource}. We evaluate pretrained models on the \textit{T0 Eval} benchmark using \texttt{transformers}\footnote{\url{https://huggingface.co/docs/transformers/index}} and \texttt{t-zero}\footnote{\url{https://github.com/bigscience-workshop/t-zero}}.

\paragraph{Pretraining and Finetuning Costs.}
Pretraining METRO-T5 in the \textit{base} setting takes 20.8 hours on 64x NVIDIA A100 (40GB) GPUs. The pretraining cost of METRO-T5 is T5 (our implementation) plus the auxiliary transformer, whose number of layers is 1/3 of the main transformer's encoder. Pretraining METRO-T5 in the \textit{base++} setting takes 159 hours on 128x NVIDIA A100 (40GB) GPUs. Pretraining METRO-T5 in the \textit{large++} setting takes 289 hours on 256x NVIDIA A100 (40GB) GPUs. In finetuning, we remove the auxiliary transformer and the RTD and CLM heads, so the finetuning cost of METRO-T5 and T5 are the same. Prompt-finetuning each \textit{base/base++} model takes about ~22 hours on 64x NVIDIA V100 (16GB) GPUs. Prompt-finetuning each \textit{large++} model takes about ~70 hours on 64x NVIDIA V100 (16GB) GPUs.

\subsection{Full Results on \textit{T0 Eval}}\label{sec:appendix_full_t0}

\begin{figure}[h]
\centering
\includegraphics[width=0.75\textwidth]{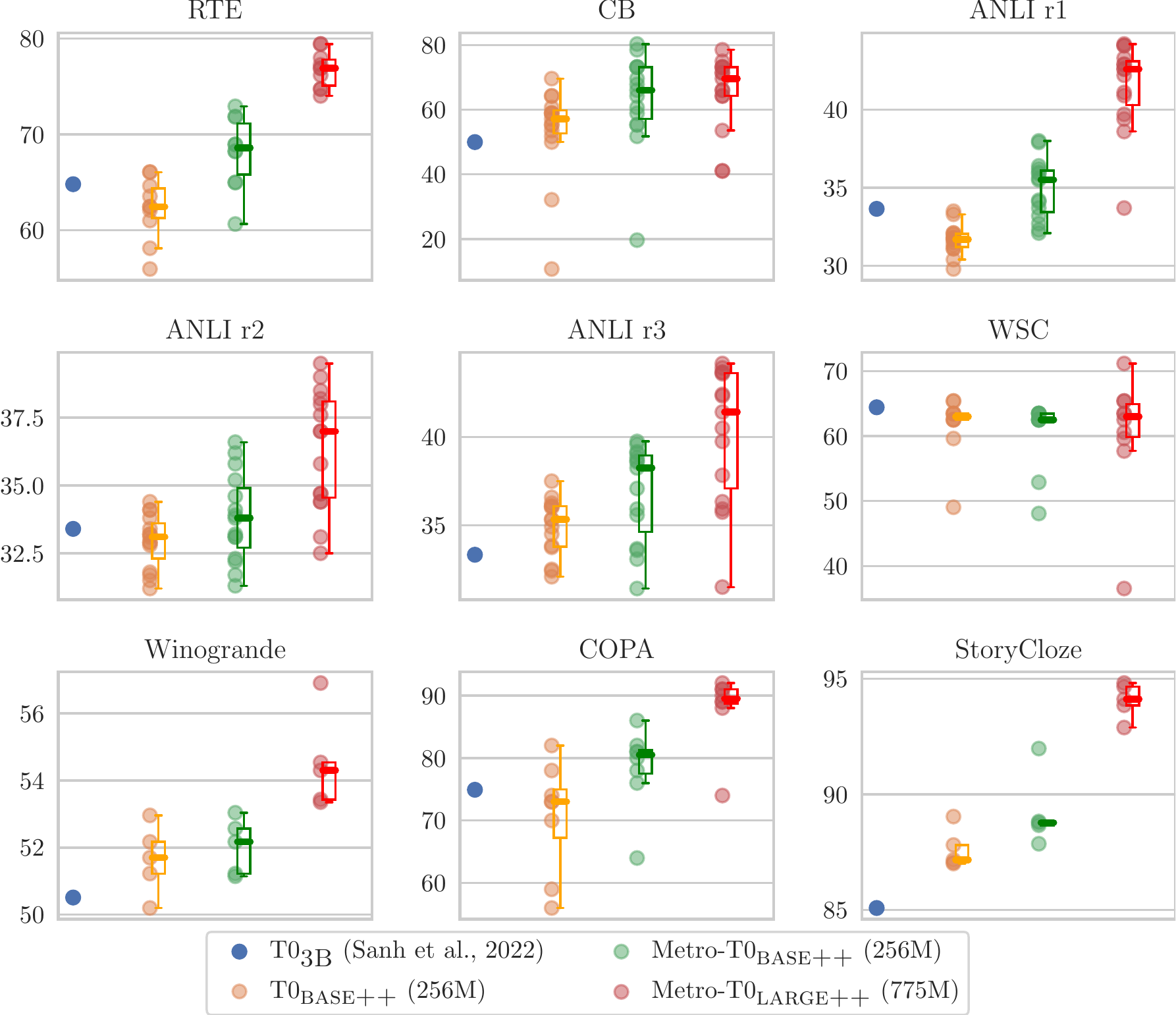}
\caption{Prompt-based learning results of METRO-T0 versus our T0 baseline and T0$_\textsc{3B}$ by \citet{sanh2022multitask} on all 9 tasks in the \textit{T0 Eval} benchmark. Each point denotes the accuracy using one prompt template, except that the median accuracy over all templates of T0$_\textsc{3B}$ is indicated by the \textcolor{tab10blue}{\textbf{blue}} point.}
\label{fig:appendix_full_t0}
\end{figure}

\Cref{fig:appendix_full_t0} results of METRO-T0 versus our T0 baseline and T0$_\textsc{3B}$ by \citet{sanh2022multitask} on all 9 tasks in the \textit{T0 Eval} benchmark. The results shows that METRO-T0$_\textsc{large++}$, having only 775M parameters, consistently outperforms T0$_\textsc{3B}$ over all tasks on the \textit{T0 Eval} benchmark.

\subsection{Evaluation on MMLU}\label{sec:appendix_evaluate_mmlu}

The prompt template used to evaluate our models MMLU is the prompt template from the \textit{AI2 Reasoning Challenge (AI2-ARC)} concatenated with 5 passages in MS MARCO~\cite{nguyen2016msmarco}. These 5 passages are selected via \textit{dense retrival} using T5-ANCE~\cite{anonymous2023augmenting,ni2021sentencet5}, which maps a query to a single vector to retrieve similar passage from the corpus. Adding densely-retrieved passages to prompts is a standard approach to enhance LM's performance on zero-shot prompting. This approach is named \textit{retrieval augmentation}. All T0 and METRO-T0 results reported in \Cref{tab:mmlu} are evaluated using this prompt template with retrieval augmentation.

On the other hand, all Flan-T5 results reported in \Cref{tab:mmlu} are numbers reported in their paper. For each model, we take the maximum score of the reported ``direct'' prompting performance and the ``chain-of-thought (CoT)'' prompting performance. Both prompt templates are not publicly available as of the time this paper is written.

As a result, \Cref{tab:mmlu} involves comparisons across multiple prompt templates. So in \Cref{tab:appendix_mmlu}, we present the performance of each model using the \textit{plain} AI2-ARC prompt template \textit{without} retrieval augmentation or CoT.

\begin{table}[h]
\centering
\small 
\begin{tabular}{lrr}
\toprule
\textbf{Model} & \textbf{Params} & \textbf{MMLU} \\
\midrule
\multicolumn{3}{l}{\textbf{AI2-ARC Prompt Template}}
\\ \midrule
T0++$_\textsc{base}$ & 226M & 31.5  \\
METRO-T0++$_\textsc{base}$ & 226M &  \textbf{31.9} \\
\midrule
Flan-T5$_\textsc{base}$~\citep{wei2022chain} & 223M & 33.8  \\
T0++$_\textsc{base++}$ & 256M & 37.8  \\
METRO-T0++$_\textsc{base++}$ & 256M & \textbf{38.9}  \\
\midrule
Flan-T5$_\textsc{large}$~\citep{wei2022chain} & 750M & 39.0  \\
T0++$_\textsc{11B}$~\citep{sanh2022multitask} & 11B & 30.9  \\
METRO-T0++$_\textsc{large++}$ & 775M &  \textbf{43.4} \\
\midrule
\multicolumn{3}{l}{\textbf{AI2-ARC Prompt Template + Retrieval Augmentation}} \\ \midrule
T0++$_\textsc{base}$ & 226M & 37.5 \\
METRO-T0++$_\textsc{base}$ & 226M & \textbf{38.3} \\
\midrule
Flan-T5$_\textsc{base}$~\citep{wei2022chain} & 223M & 40.4  \\
T0++$_\textsc{base++}$ & 256M & 41.7 \\
METRO-T0++$_\textsc{base++}$ & 256M & \textbf{42.7} \\
\midrule
Flan-T5$_\textsc{large}$~\citep{wei2022chain} & 750M & 41.4  \\
T0++$_\textsc{11B}$~\citep{sanh2022multitask} & 11B & 35.6 \\
METRO-T0++$_\textsc{large++}$ & 775M & \textbf{48.0} \\
\midrule
\multicolumn{3}{l}{\textbf{Reported numbers by \citet{chung2210scaling}}} \\
\midrule
Flan-T5$_\textsc{base}$~\citep{wei2022chain} & 223M & 35.9 \\
\midrule
GPT-3$_\textsc{175B}$~\citep{brown2020language} & 175B & 43.9 \\
Flan-T5$_\textsc{large}$~\citep{wei2022chain} & 750M & 45.1 \\

\bottomrule
\end{tabular}
\caption{Full prompt learning results on the \textit{MMLU} dataset in three setups. All reported results use accuracy averaged over 57 subtasks as their metric.}
\label{tab:appendix_mmlu}
\end{table}

The result in \Cref{tab:appendix_mmlu} shows that METRO-T0++ still consistently outperforms the T0 baseline and similar-sized Flan-T5 models when they are evaluated using the same prompt template.
\subsection{Example of the Challenge of Ill-Formed Target}\label{sec:appendix_ill_form}

In our discussion about ``decoding target'' in\Cref{sec:method_metro}, we claim that \textit{``masked tokens only''} is an \textit{ill-formed target} for the CLM objective in METRO-style pretraining of T5. This section shows a concrete example where such ill-formed target leads to ambiguities.

\begin{table}[htbp]
\small
\centering
\caption{
An example where ill-formed target leads to ambiguities. Each number denotes a distinct subword token. \texttt{M} denotes the special token \texttt{[MASK]}. In ``Auxiliary Model Prediction'', a token shown in \textcolor{teal}{green} denotes a correct prediction, where a token shown in \textcolor{red}{red} denotes a wrong prediction.
}\label{tab:appendix_ill_form}
\begin{tabular}{ll}
\toprule
\textbf{Sentence} & \texttt{1 2 3 4 5} \\
\midrule
\textbf{Auxiliary Model Input 1} & \texttt{1 M M M 5} \\
Auxiliary Model Prediction & \texttt{\ \ \textcolor{teal}{2} \textcolor{red}{6} \textcolor{teal}{4}  }\\
Main Model Input & \texttt{1 2 6 4 5} \\
Main Model Target & \texttt{2 3 4} \\
\midrule
\textbf{Auxiliary Model Input 2} & \texttt{1 2 M M 5} \\
Auxiliary Model Prediction & \texttt{\ \ \ \ \textcolor{red}{6} \textcolor{teal}{4}  }\\
Main Model Input & \texttt{1 2 6 4 5} \\
Main Model Target & \texttt{3 4} \\
\bottomrule
\end{tabular}
\end{table}

In \Cref{tab:appendix_ill_form}, the original sentence is ``\texttt{1 2 3 4 5}''. Using different random samples of masked positions, we can derive two masked sequences as the input of the auxiliary model: ``\texttt{1 M M M 5}'' and ``\texttt{1 2 M M 5}''. The difference is whether ``\texttt{2}'' is masked or not. So the target for the decoder corrective LM objective will be ``\texttt{2 3 4}'' and ``\texttt{3 4}'' respectively. After we have the masked input, the auxiliary model, which is a \textit{masked language model (MLM)}, tries to fill masked positions with predicted tokens ``\texttt{2 6 4}'' and ``\texttt{6 4}'' respectively. The resulting main model input is ``\texttt{1 2 6 4 5}'' for both cases, but the target is ``\texttt{2 3 4}'' for case 1 and ``\texttt{3 4}'' for case 2. This is an ambiguity where the main model is unsure where it should begin to generate predictions: ``\texttt{2}'' or ``\texttt{3}''.
% \section{Formal Definition of Parameter Sensitivity}\label{sec:appendix_sensitivity}

% The \textit{sensitivity} of a parameter is defined in \Cref{eqn:sensitivity}. $\theta$ denotes the parameter vector and $\mathcal{L}$ denotes the loss function. $\theta_{-j}$ denotes the parameter vector $\theta$ with its $j$-th entry set to zero. The approximation is derived from the first-order Taylor expansion of $\mathcal{L}$ at $\theta$. Therefore, the sensitivity of the $j$-th parameter, denoted by $I_j$, approximates the change in the loss magnitude when this parameter is completely zeroed-out~\citep{lecun1989optimal}.

% \begin{equation}
% I_j = |\theta_{-j}^T\nabla_\theta \mathcal{L}(\theta)| \approx |\mathcal{L}(\theta) - \mathcal{L}(\theta - \theta_{-j})| 
% \label{eqn:sensitivity}
% \end{equation}

% \citet{liang2022noparameter} shows that parameter sensitivity is a reliable indicator of redundancy in pretrained language models. Specifically, parameters with low sensitivity can be safely pruned with only a marginal impact on the LM's downstream performance, and an LM with lower, more concentrated sensitivity is more sufficiently trained and generalizes better.

\end{document}